\documentclass[letterpaper, 10 pt, conference]{ieeeconf} 

\let\labelindent\relax

\IEEEoverridecommandlockouts                   

\title{\LARGE \bf
Characterizing Structured versus Unstructured Environments based on Pedestrians' and Vehicles' Motion Trajectories*
}

\author{Mahsa Golchoubian$^{1\dag}$, Moojan Ghafurian$^{2}$, Nasser Lashgarian Azad$^{1}$, Kerstin Dautenhahn$^{1,2}$
\thanks{* This research was undertaken, in part, thanks to funding from the Canada 150 Research Chairs Program and NSERC.}
\thanks{\dag Corresponding Author: Mahsa Golchoubian {\tt\small mahsa.golchoubian@uwaterloo.ca}}
\thanks{$^{1}$Department of Systems Design Engineering, University of Waterloo, Canada}
\thanks{$^{2}$Department of Electrical and Computer Engineering, University of Waterloo, Canada}%
}

\usepackage{graphicx}
\usepackage{subfigure}
\usepackage{tablefootnote}
\usepackage{enumitem}
\usepackage{dblfloatfix}  

 \usepackage[pscoord]{eso-pic}
\newcommand{\placetextbox}[3]{
 \setbox0=\hbox{#3}
 \AddToShipoutPictureFG*{ \put(\LenToUnit{#1\paperwidth},\LenToUnit{#2\paperheight}){\vtop{{\null}\makebox[0pt][c]{#3}}}
 }
 }

  \placetextbox{.5}{0.055}{\footnotesize{© 2022 IEEE. Personal use of this material is permitted. Permission from IEEE must be obtained for all other uses, in any current or future media,}} 
  \placetextbox{.5}{0.045}{\footnotesize{including reprinting/republishing this material for advertising or promotional purposes, creating new collective works, for resale or redistribution}}
  \placetextbox{.5}{0.035}{\footnotesize{to servers or lists, or reuse of any copyrighted component of this work in other works.}}

\begin{document}

\maketitle
\thispagestyle{empty}
\pagestyle{empty}

\begin{abstract}

Trajectory behaviours of pedestrians and vehicles operating close to each other can be different in unstructured compared to structured environments. These differences in the motion behaviour are valuable to be considered in the trajectory prediction algorithm of an autonomous vehicle. However, the available datasets on pedestrians’ and vehicles’ trajectories that are commonly used as benchmarks for trajectory prediction have not been classified based on the nature of their environment. On the other hand, the definitions provided for unstructured and structured environments are rather qualitative and hard to be used for justifying the type of a given environment. In this paper, we have compared different existing datasets based on a couple of extracted trajectory features, such as mean speed and trajectory variability. Through K-means clustering and generalized linear models, we propose more quantitative measures for distinguishing the two different types of environments. Our results show that features such as trajectory variability, stop fraction and density of pedestrians are different among the two environmental types and can be used to classify the existing datasets.

\end{abstract}

\section{Introduction}

Many researchers in the area of motion prediction and planning are using the terms structured and unstructured environments for addressing two distinct application areas for their automated systems \cite{chen2021spatial,su2019potential,cai2020context,guastella2021learning, procopio2009learning}. Different definitions have been proposed in the literature for classifying these environments into structured and unstructured.  In one type of definition that is focused on vehicular traffic environments, the distinction is made based on the existence of lane marks and strict traffic rules (e.g., see~\cite{jyothi2019driver,kerscher2018intention,kolski2006autonomous,ibrahim2013collision}). Within this definition, a structured environment is believed to be a place with clear lane marks and road dividers, and where traffic participants adhere to strict traffic rules~\cite{jyothi2019driver}. On the other hand, in an unstructured environment, such marked lanes or road dividers do not exist and no strict traffic rules are followed by drivers and pedestrians. In unstructured environments, all traffic participants are required to pay more attention due to the reduced traffic rules~\cite{jyothi2019driver}.  Pedestrianized spaces and parking lots are considered as unstructured environments under this definition~\cite{jyothi2019driver,kerscher2018intention}.

In another definition, the coherency of the movement pattern has been introduced as the aspect that distinguishes structured from unstructured environments~\cite{ozturk2010detecting,jimenez2019pedestrian,rodriguez2009tracking}. In this definition, an environments is called structured when a main motion track can be identified which is usually caused by a condition in the environment, such as the pedestrian crosswalks on roads. In contrast, in unstructured environments, pedestrians can move freely in different directions~\cite{ozturk2010detecting}. In this definition, unlike the previous one, a pedestrianized environment can be either structured or unstructured based on the existing motion patterns.

Although there are some overlaps between these two definitions, there are also some differences that make it hard to specify whether an environment is structured or unstructured solely based on these qualitative definitions.

Despite the differences in the two definitions provided above, it is agreed in the literature that the behaviours of traffic participants in these two types of environments are different. This becomes important for an autonomous vehicle that needs to predict the future motions of other traffic participants~\cite{leon2021review,abnili2019short}, as this prediction will differ based on the type of the environment in which the vehicle is operating.

Many data-driven trajectory prediction models have been proposed for the use of autonomous transportation systems.
These models are usually trained on datasets that contain real-world trajectories of pedestrians and vehicles in an environment. Depending on the type of the environment these datasets have been captured from, the trained predictive model will also learn the motion behaviour of traffic participants within the same type of environment. Paying attention to this point is crucial when it comes to generalizing the trained motion predictor to other similar environments. 

Therefore, while it is valuable to classify the datasets based on the type of the environment (i.e., unstructured versus structured), it is hard to do this classification based on the different qualitative definitions provided in the literature. Therefore, the present work investigates how to quantitatively define the type of environments through evaluating quantitative features for classifying environments to unstructured and structured, by focusing on the traffic participants’ behaviours embedded in the trajectories. We use two methods, (a) clustering, to see if the existing datasets can be divided into two clusters showing structured vs. non-structured environments, and (b) regression to investigate if our assumptions about the effect of features on identifying the type of environment is hold. Based on the existing literature, we assume that traffic participants have similar behaviours in each type of environment, and based on this similarity we will suggest a relationship between the extracted quantitative features and the structured/unstructured nature of the environment. To the best of our knowledge, this is the first work that proposes a quantitative characterization for a structured vs. unstructured environment based on the trajectory data.

As we are interested in classifying environments that include both pedestrians and vehicles into structured and unstructured environments, we only analyze the datasets that have both of these agents present.

In this paper, we address two main questions: (1) How can we quantify the definition of structured and unstructured environments that include both pedestrians and vehicles according to trajectory data. (2) How the existing datasets on agents' (i.e., pedestrians and vehicles) trajectories will get categorized based on these quantitative measures. 
Considering these two research questions, our contribution is two-fold: (1) We suggest practical quantitative features that can be used for classifying environments as structured and unstructured. (2) We group some of the well-known datasets in two classes based on these features.  

Our proposed quantitative features can be used to identify the type of a new target environment, in order to use existing pedestrian trajectory datasets classified within the same environmental type for training a trajectory prediction algorithm. This might be particularly beneficial for designing autonomous vehicles operating in different environments, so that they can improve the trajectory prediction of different traffic participants by considering their behaviour in that specific environmental type.

\section{Related Work}

Pedestrian trajectory prediction is an active field of research with different application areas in autonomous driving and social robotics \cite{rudenko2020human}. Various approaches have been used for this trajectory prediction from physics-based models such as social force model \cite{rudenko2018joint,rinke2017multi,predhumeau2021agent} to data-driven models especially deep learning ones \cite{salzmann2020trajectron++,chandra2019traphic,cheng2020mcenet}. Corresponding to this line of research, many real-world datasets have been collected from different environments to be used as a benchmark for these trajectory prediction algorithms. While the nature of the environment and vehicles' presence can have a considerable influence on the trajectory profile of pedestrians, no clear classification has been made between the different datasets based on quantitative measures such as the pedestrians’ and vehicles' trajectory features. 

Different datasets have been compared in terms of the complexity of the human trajectories for the prediction task by introducing some indicators in \cite{amirian2020opentraj}. However, the focus of this comparison was only on pedestrians’ behaviour and therefore mostly included datasets in which pedestrians are the only agent present without being influenced by vehicles. 

Robicquet et al. have detected different classes of navigation styles for pedestrians in a dataset based on the different ways pedestrians handle their interaction in a collision avoidance situation \cite{robicquet2016learning}. But it is still unclear how these navigation styles can also be affected by the type of the environment. 

Furthermore, traffic participants’ spatial and movement behaviour in different shared space layouts is studied in \cite{batista2022investigating}, considering both vehicles and pedestrians. But the number of behavioural features studied was limited to mean speed, number of interactions, and yield ratio.

Also, in another study focusing on pedestrians, pedestrians’ behaviour in a shopping mall was classified into three categories of “walking straight”,  “finding a way”, and “walking around” in \cite{okamoto2011classification}, showing that roaming around is a common trajectory behaviour that can be seen in a shopping mall environment.
However, the analysis was restricted to one dataset with only pedestrians moving in a mall. Nevertheless, we get inspiration from some of the behavioural measures used in these papers (e.g., path efficiency \cite{amirian2020opentraj}, stop ratio \cite{okamoto2011classification}) for building the quantitative features that can describe unstructured vs. structured environments. 

\section{Datasets}

There are a few datasets that include both pedestrians and vehicles which are analyzed in this paper, discussed below.

\begin{itemize}[leftmargin=0cm]
    \item[] \textbf{Stanford Drone Dataset (SDD)} is a dataset that has been captured from the campus of Stanford university and consists of eight different scenes \cite{robicquet2016learning}. The dataset includes trajectories of around 19k agents in different classes from pedestrians, bicyclists, skateboarders to cars, carts, and buses. In this paper, we have analyzed the trajectories of only pedestrians, cars and carts within four scenes of this dataset (hyang, coupa, nexus, death circle) that include a relatively higher percentage of car/cart compared to the other scenes. In this dataset, we filtered out the annotations with ``lost" value of 1 which indicates annotations outside the view screen.
    \item[] \textbf{inD dataset} contains trajectories of more than 11K traffic participants including pedestrians, vehicles, and bicycles \cite{bock2020ind}. The dataset was recorded from four different German intersections with two of these intersections being used in the analysis of the current article.  
    \item[] \textbf{DUT dataset} contains trajectories of pedestrians and vehicles interacting with each other in two locations on the campus of Dalian University of Technology in China with close to 2K pedestrian and 65 vehicle trajectory data \cite{yang2019top}. One of the locations is an intersection without a traffic light. The other location is a shared space near a roundabout where agents can move freely. 
    \item[] \textbf{Hamburg Bergedorf Station (HBS)} dataset is captured from a shared space street in Germany near a busy railway station \cite{pascucci2017discrete}. It includes trajectory data of 1115 pedestrians, 331 cars and 29 cyclists.
    \item[] \textbf{HC dataset} was captured from a shared street in a university campus in Germany \cite{cheng2019pedestrian}. It contains trajectory data of 282 pedestrians, 15 vehicles and 40 cyclists.
\end{itemize}

All the above datasets have been captured from stationary cameras or drones. In each of these datasets, the position track of different agents is provided frame by frame. These positions were all transformed to a world cartesian coordinate before being used for analysis. 

Three of the above datasets (i.e. HBS, HC and the roundabout scene of the DUT dataset) are collected from shared spaces environments which are defined as urban environments where the curbs, road surface markings and traffic signs have been removed to encourage different traffic participants to share the same space and negotiate the priority based on social rules \cite{predhumeau2021pedestrian}. 
However, it is not clear whether shared spaces should be categorized as unstructured or structured environments according to the motion behaviour. This would be further investigated during our analysis.

The secondary use of the above datasets for this study has received ethics clearance from the University of Waterloo Research Ethics Committee.

\section{Method} 

For each dataset and for all of its different scenes, we extracted thirteen features from the traffic participants’ trajectories that could be helpful in distinguishing a structured environment from an unstructured one.
We have categorized our extracted features into three sets of pedestrian motion features, vehicle motion features, and pedestrian-vehicle interaction features that are described in order.

\subsection{Extracted trajectory features}

\textbf{Pedestrian motion features:} The features that we extracted from each individual pedestrian’s trajectory were: (a) mean speed of pedestrian, (b) stop fraction of pedestrian, (c) variability of pedestrian's trajectory, (d) path efficiency of pedestrian, (e) pedestrians' path orientation entropy, (f) density of pedestrians, and (g) density of standing pedestrian.

The first three features are related to the speed profile of the pedestrian. We calculated the velocity vector from the pedestrian’s position vector in consecutive frames knowing the frame rate through which the dataset was captured. For the mean speed feature, we calculated the average speed of the pedestrian during its whole trajectory. For the stop fraction, we computed the percentage of time the pedestrian is in a stop along its whole trajectory. We have considered a pedestrian to be in a stopping position when their speed gets lower than 0.5 m/s \cite{okamoto2011classification}.
Variability of trajectory is defined as the trace of the covariance matrix of the speed \cite{okamoto2011classification} and is calculated according to the equation below:

\begin{equation}
\label{eq:Variability} 
    tr(cov(\overrightarrow{V})) = \frac{1}{n} \sum_{t} (v_t - \bar{v})^2
\end{equation}

\begin{figure}[b!]
  \centering
  \fbox{\includegraphics[height=2.5cm]{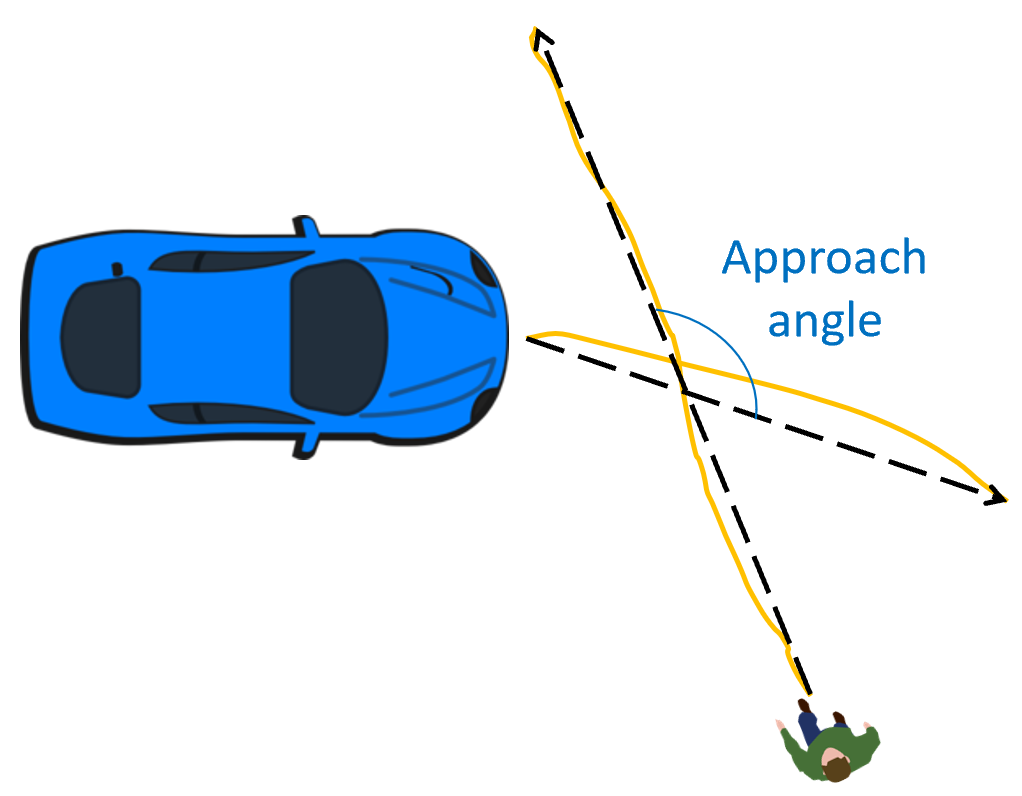}}
  \caption{Approach angle definition in a pedestrian-vehicle interaction}
  \label{fig:ApA}
\end{figure}

\begin{figure*}[b!]
\centering
\subfigure[Pedestrian mean speed]{
\fbox{\includegraphics[height=4cm]{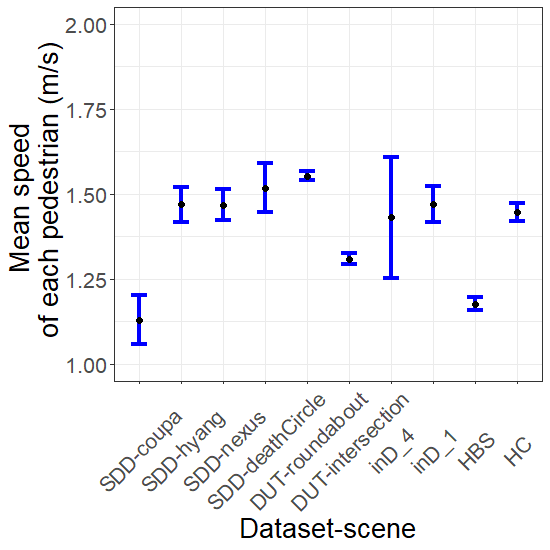}}  
\label{fig:pedSp}
}
\subfigure[Pedestrian stop fraction]{
\fbox{\includegraphics[height=4cm]{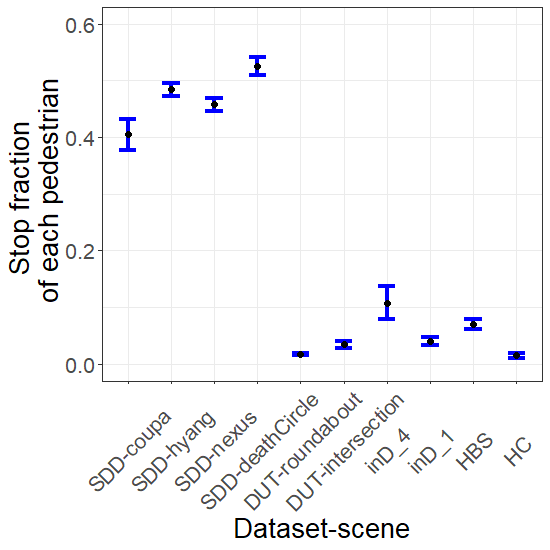}}
\label{fig:pedSt}
}
\subfigure[Variability of pedestrian's trajectory]{
\fbox{\includegraphics[height=4cm]{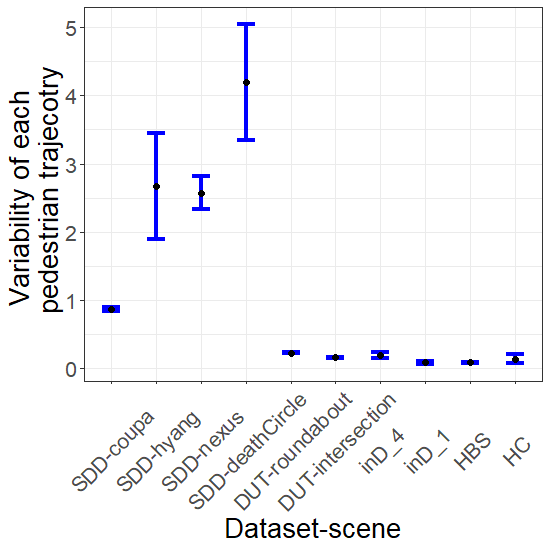}}
\label{fig:pedVa}
}
\subfigure[Pedestrian path efficiency]{
\fbox{\includegraphics[height=4cm]{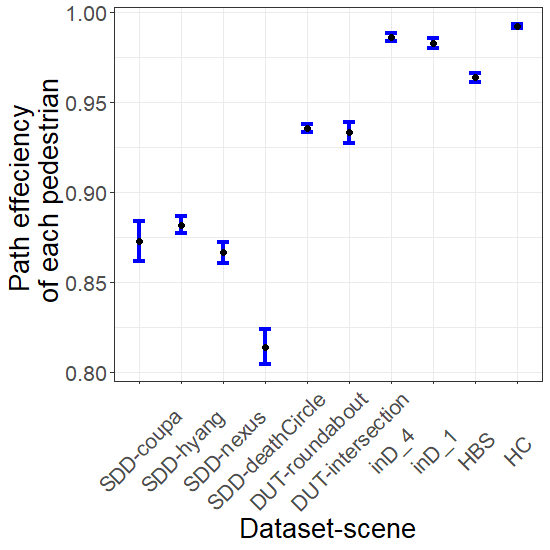}}
\label{fig:pedEf}
}\hspace{1cm} 
\subfigure[Entropy of path orientation]{
\fbox{\includegraphics[height=4cm]{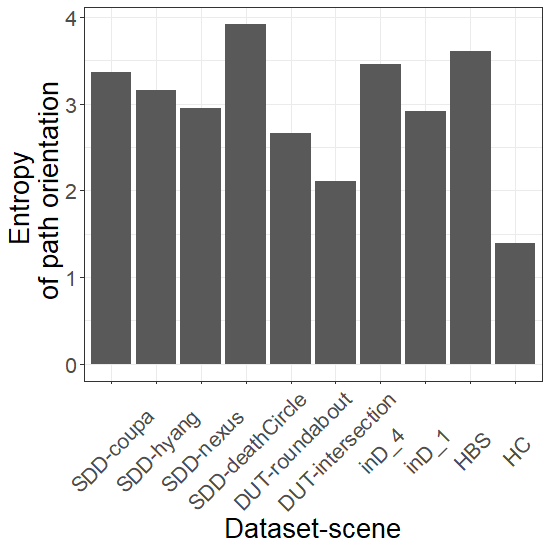}}
\label{fig:pedPO}
}
\subfigure[Pedestrian Density]{
\fbox{\includegraphics[height=4cm]{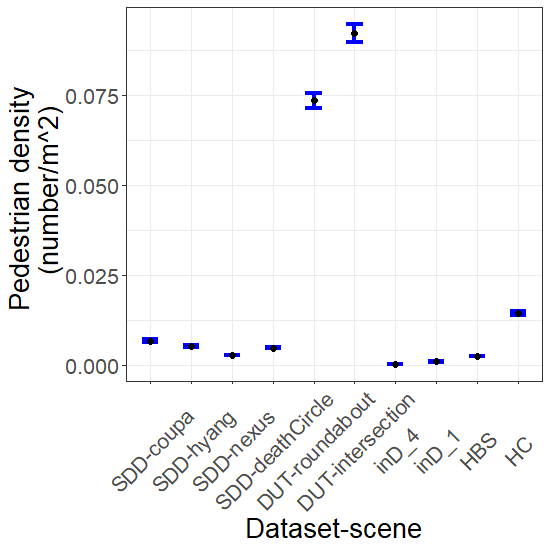}}
\label{fig:pedDe}
}
\subfigure[Standing Pedestrian Density]{
\fbox{\includegraphics[height=4cm]{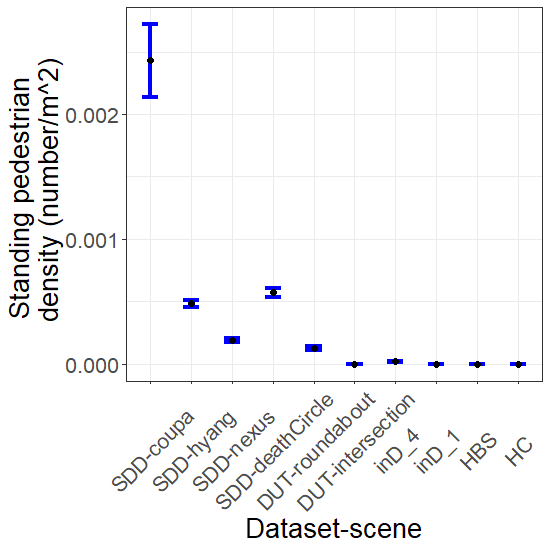}}
\label{fig:stDen}
}
\caption{The value of features extracted from pedestrians' trajectories in each dataset with 95 \% confidence interval}
\label{fig:PedFe}
\vspace{-12px}
\end{figure*}

In this equation, $\overrightarrow{V}$ is a vector of speeds in different time steps t, ($\overrightarrow{V} = [v_1, v_2, ..., v_n]$), $\bar{v}$ is the average speed and $n$ is the total number of time steps that the pedestrian is present in the scene. 

We calculated path efficiency as the ratio of the distance between the endpoints of the trajectory over the trajectory length and indicates how close the path is to a straight line \cite{amirian2020opentraj}.
Some of the datasets have a wider perspective of the capturing scene and therefore in those scenes, the pedestrians' trajectories can deviate from a straight line due to some geometrical constraints in the scene such as the buildings. To compensate for that in the calculation of the path efficiency, we divide the trajectories into smaller sections each lasting 4.8 seconds and the path efficiency is calculated using the cumulative value of the smaller sections of the trajectory within the formulation. These smaller pieces of the trajectory that last 4.8 seconds are called trajlet in \cite{amirian2020opentraj} and are commonly used in trajectory prediction as the observation interval of each agents’ trajectory for future position prediction.

For the path orientation feature, we calculate the angle of the straight line connecting the start and endpoint of each trajlet relative to a global horizon. However, the exact number of the path orientation is not important and cannot be compared between the different datasets due to the inconsistency of the orientation of their global Cartesian frame. In fact, what is important is the diversity of the path orientation within each environment that could represent the pedestrians' level of freedom in selecting their movement direction. For measuring this diversity, we calculate the entropy of the path orientation for each dataset and keep it as a feature.

All the above features are calculated for the pedestrians that are not stationary throughout the whole dataset. Stationary pedestrians are considered as those with a stop fraction of more than 90\% and are excluded from the above feature calculations as those features are not meaningful for stationary pedestrians and the calculations are prone to high error for pedestrians that have a speed of more than 0.5 m/s only at 10\% of their whole trajectory. However, for calculating the density of the pedestrians, all the moving and standing pedestrians are being considered and their total number is divided by the area of the scene from which the dataset is being captured. As the stationary pedestrians were ignored in the first 5 features, we added a last feature that specifies the density of standing pedestrians in the dataset. These two density features are calculated for each pedestrian as the average density while the pedestrian is present in the scene since the density of the crowd could affect each individual's motion behaviour. 

\textbf{Vehicle motion features:} From the vehicle trajectory data we extracted:  (a)  mean speed of the vehicle, (b) stop fraction of the vehicle, and (c) variability of vehicle's trajectory.

We defined and calculated these features in the same way as in the pedestrian features. The only difference is in the stop fraction feature, where a vehicle is considered to be in a stop when its speed is less than 1 m/s. Same as in the pedestrians feature, here we also exclude vehicles that have a means speed of less than 0.5 m/s throughout their whole presence which is used as an indication for a parked vehicle.

\textbf{Pedestrian-vehicle interaction features:} The way pedestrians and vehicles interact in a structured vs. unstructured environment can also be very different. In an unstructured environment as there are no pre-specified paths or lane marks present \cite{kerscher2018intention}, pedestrians and vehicles can approach one another from different directions and since no strict rules apply in unstructured environments~\cite{jyothi2019driver}, different behaviours can be observed in terms of which class of agent (pedestrians or vehicles) gets the priority when their paths cross one another. Therefore, (a) entropy of approach angle, (b) priority (percentage of pedestrian priority), and (c) ratio of vehicles to pedestrians were the three features that we defined for capturing these differences.

For extracting these behaviours as interaction features, we first focus on those pairs of pedestrian-vehicle that get closer to each other than a threshold that we have selected to be 4 meters here. Based on the speed and size of the vehicles, this four meters was a rational distance choice below which calculating approach angle was justifiable.

For each pair of pedestrian-vehicle, we then calculate their approach angle as the angle between the straight line connecting the endpoints of each of the two agent’s trajectories while they are in interaction~(Fig. \ref{fig:ApA}). We will then calculate the entropy of the approach angle distribution within each dataset as a measure of the approach angle diversity.
As another feature we calculate the ratio of pedestrians gaining priority over vehicles to the total number of times a path crossing situation happened between a vehicle and pedestrians during their interaction. We defined priority in terms of who got to first cross the intersection point of the two crossing trajectories.
We also include the ratio of vehicles to pedestrians in each frame as a feature, with the average of this ratio being calculated over all the frames of the dataset.

\begin{figure*}
\centering
\subfigure[Vehicle mean speed]{
\fbox{\includegraphics[height=3.9cm]{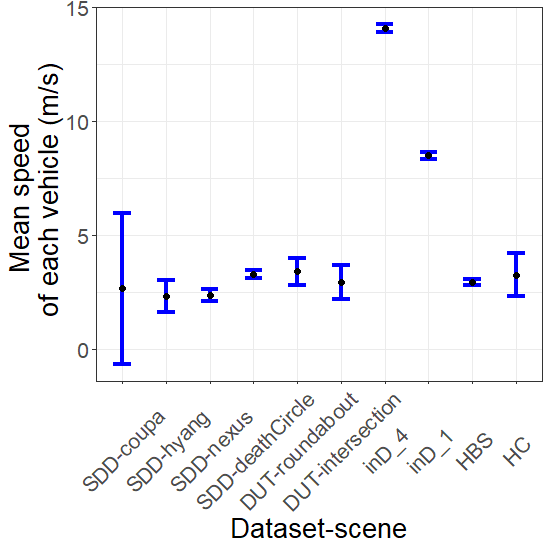}}  
\label{fig:VehSp}
}
\subfigure[Vehicle stop fraction]{
\fbox{\includegraphics[height=3.9cm]{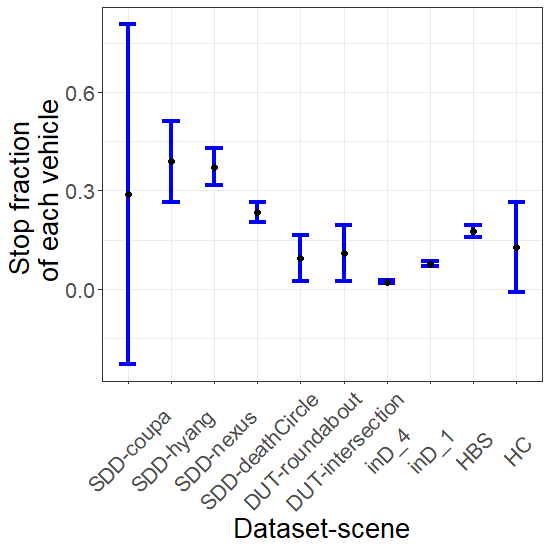}}
\label{fig:VehSt}
}
\subfigure[Variability of Vehicle 's trajectory]{
\fbox{\includegraphics[height=3.9cm]{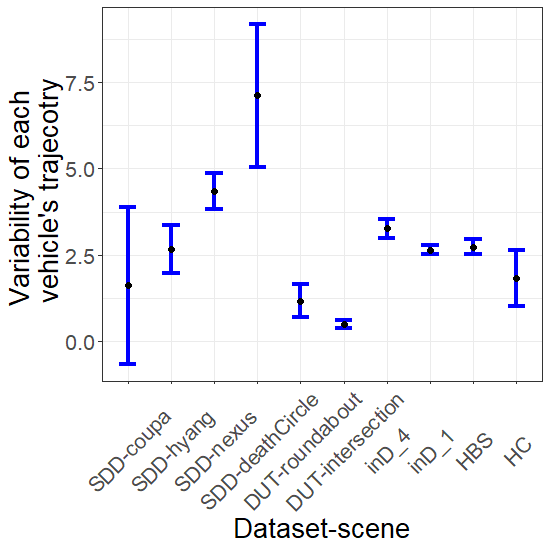}}
\label{fig:VehVa}
}
\caption{The value of features extracted from vehicles' trajectories in each dataset with 95 \% confidence interval}
\label{fig:VehFe}
\vspace{-12px}
\end{figure*}

\begin{figure*}[t!]
\centering
\subfigure[Entropy of approach angle]{
\fbox{\includegraphics[height=4cm]{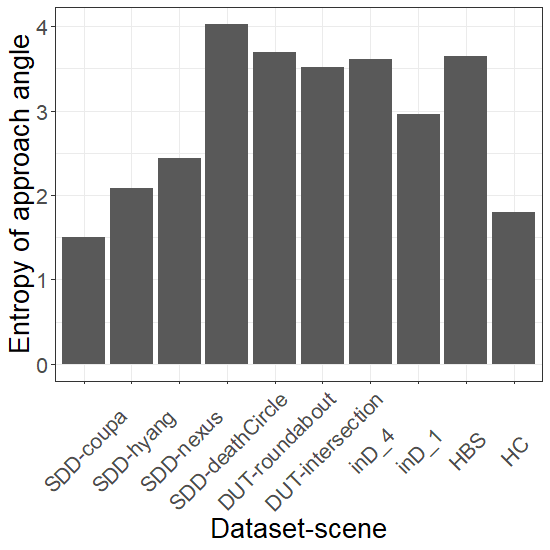}}  
\label{fig:Appr}
}
\subfigure[Priority in path crossing]{
\fbox{\includegraphics[height=4cm]{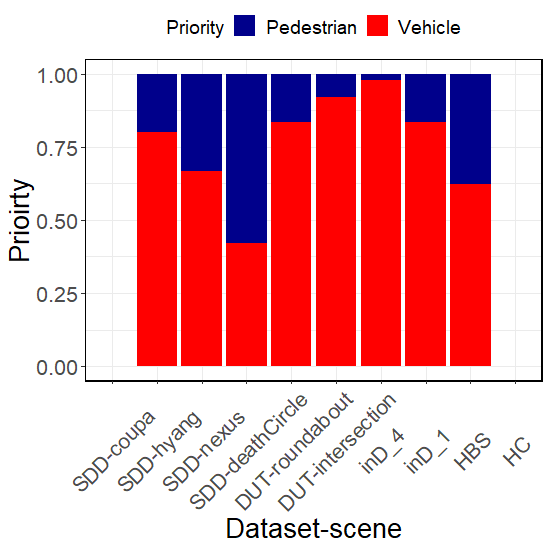}}
\label{fig:Prio}
}
\subfigure[Ratio of vehicles to pedestrians]{
\fbox{\includegraphics[height=4cm]{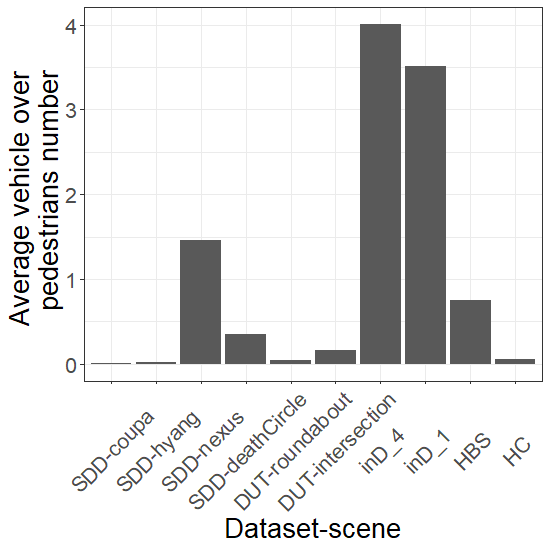}}
\label{fig:V2P}
}
\caption{The value of features extracted from pedestrian-vehicle interaction in each data}
\label{fig:IntFe}
\end{figure*}

\subsection{Data Analysis Approach} 

Having the above-extracted features for each dataset, we apply k-means clustering to find two clusters that could represent structured vs. unstructured environments for the data in the combined datasets. By using the k-means clustering technique we have the flexibility of choosing the number of clusters to be two, for representing the two types of environments that we wanted to distinguish between. To have all features together, we used the average of the vehicle features for each dataset and added them to the interaction and pedestrians features.

We have eliminated the outliers by using the $1.5 \times IQR$ (Interquartile range) method. Therefore, the data points that are $1.5 \times IQR$ below the 25\% percentile or above the 75\% percentile are excluded. 

As the next step we evaluate each feature for its significant effect on the clustering result. For that we fit a generalized linear model to the data, using the pedestrian features as the fixed effect and the label of the environment from the clustering result as the output. This was done using the glm function in RStudio. The result of these analyses is presented in the next section.

\section{Results and Discussion}

The mean and 95\% confidence intervals for each feature are shown in Figs.~\ref{fig:PedFe},~\ref{fig:VehFe}, and~\ref{fig:IntFe}. Note that the different scenes of each dataset are plotted separately. According to Fig.~\ref{fig:PedFe}, pedestrians in the SDD dataset have higher stop fraction and trajectory variability with a lower path efficiency compared to the other datasets. In Fig.~\ref{fig:PedFe}f, it is observed that the DUT dataset has the highest pedestrian density as it has been captured from a small congested area, but the density of standing pedestrians is higher in SDD according to Fig~\ref{fig:PedFe}g. The highest vehicle mean speed is from the inD dataset as it is being captured from a typical road environment (Fig~\ref{fig:VehFe}). The inD dataset also has the highest ration of vehicles to pedestrians among all the datasets (Fig.~\ref{fig:IntFe}c) and also the highest percentage of vehicle priority (Fig.~\ref{fig:IntFe}b).

\begin{figure}
  \centering
  \includegraphics[width=0.55\linewidth]{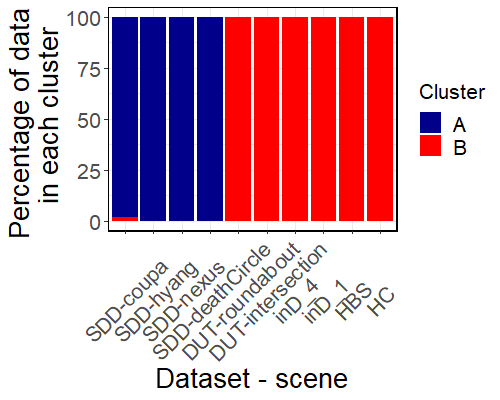}
  \caption{Datasets clustered in two groups based on the defined features. For each dataset the portion of its data points in each cluster is reported.}
  \label{fig:Clus}
\end{figure}

The clustering results for each dataset based on these thirteen defined trajectory features is shown in Fig.~\ref{fig:Clus}. This figure shows the percentage of the dataset that goes into each of the two clusters, A and B. For majority of the dataset, all the data points stand unanimously within one cluster.

According to this clustering result, cluster A which contains the four different scenes of the SDD dataset represents trajectory behaviours that happen in a campus environment and mainly off the road. All the other datasets fall into the second cluster, cluster B, which contains mostly road environments having either a conventional structure with designated operational areas for pedestrians and vehicles or having the design of a shared street. 

\begin{figure*}
  \centering
  \includegraphics[width=0.7\linewidth]{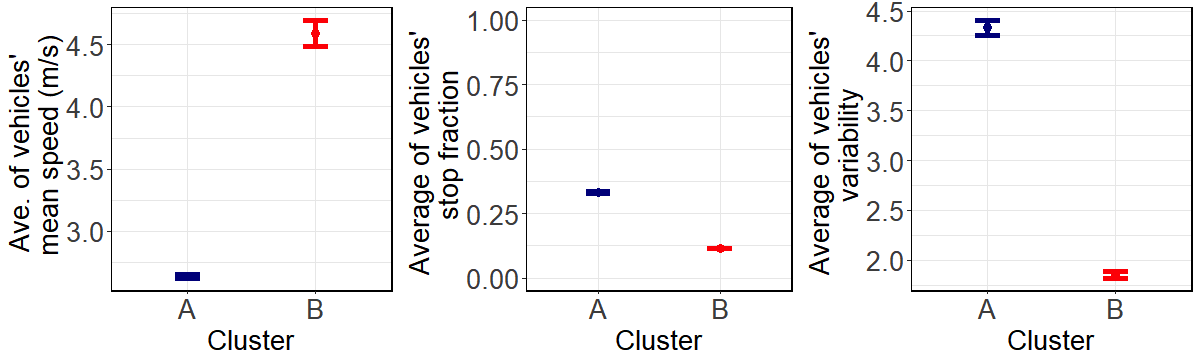}
  \caption{The mean of the vehicle features within each cluster along with the 95\% confidence intervals}
  \label{fig:VbyC}
\end{figure*}

\begin{figure*}
  \centering
  \includegraphics[width=0.7\linewidth]{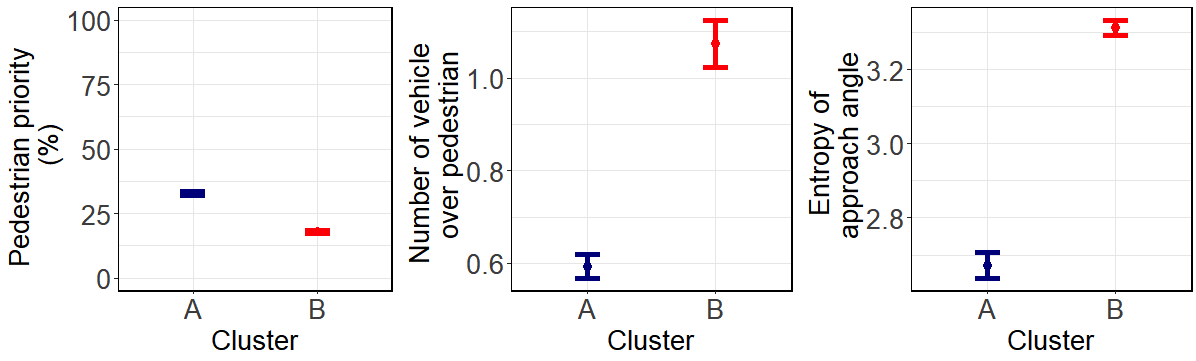}
  \caption{The mean of the interaction features within each cluster along with the 95\% confidence intervals}
  \label{fig:IbyC}
\end{figure*}

\begin{figure*}
\centering
\subfigure{
\includegraphics[width=0.9\linewidth]{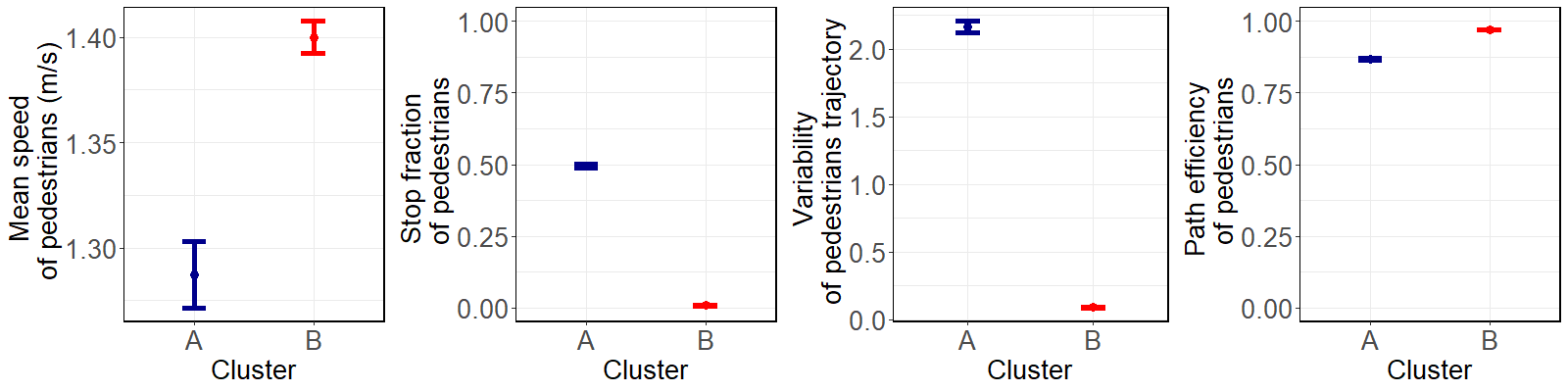}  
\label{fig:PbyC1}
}\hspace{1cm} 
\subfigure{
\includegraphics[width=0.7\linewidth]{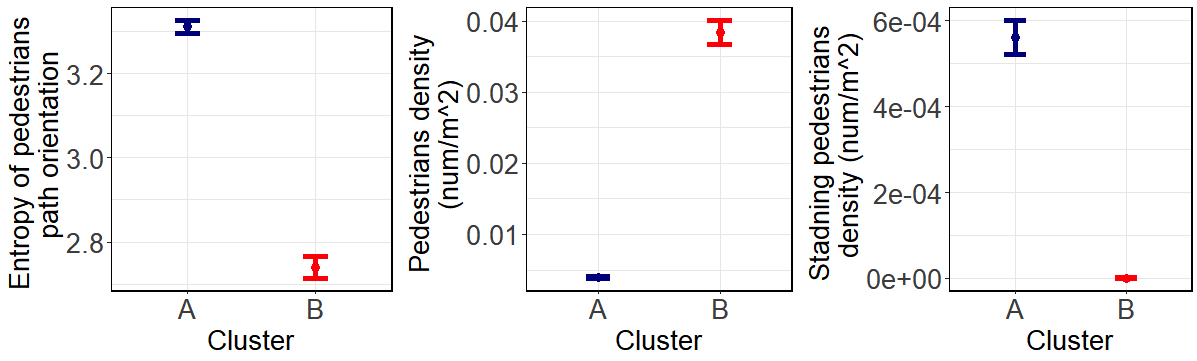}
\label{fig:PbyC2}
}
\caption{The mean of the pedestrian features within each cluster along with the 95\% confidence intervals}
\label{fig:PbyC}
\vspace{-12px}
\end{figure*}

By assigning all the points in a dataset to a label that represents the label of the majority of the points in that dataset from the k-means clustering, we have then compared the values of each feature between the two clusters in Figs. \ref{fig:VbyC}-\ref{fig:PbyC}. In these figures, the mean value of each feature is shown with the 95\% confidence interval. The statistical analysis using linear mixed-effects models (LMMs) \cite{bates2014fitting} for each feature separately while considering the dataset as a random effect shows that pedestrians' stop fraction ($se=0.005, t=-91.06, p<0.001$), the variability of pedestrians' trajectory ($se=0.069, t=-30.11, p<0.001$), and the pedestrians' path efficiency ($se=0.023, t=4.731, p<0.05$) are significantly different among the two clusters. The path efficiency is significantly lower in cluster A compared to cluster B. In addition, the density of standing pedestrians has significantly different values in the two clusters ($se=1.82e-5, t=-30.86, p<0.001$). Among the vehicle-related features, the stop fraction of the vehicle ($se=0.051, t=-4.157, p<0.05$) is significantly different between the two clusters.

\begin{table}[b!]
\centering
\footnotesize
\caption{The generalized linear model leading to best model fit according to the AIC criteria (AIC=72.83). Ped: Pedestrian.} \label{tab:glm}
\resizebox{\columnwidth}{!}{\begin{tabular}{|l|c|c|c|l|}
\hline
\textbf{Fixed effects} & \textbf{Estimate} & \textbf{Std. Error} & \textbf{z value} & \textbf{Pr(\boldmath${>|z|}$)}\\
\hline
Ped mean speed & 2.897 & 1.129 & 2.567 & 0.01 (*)\\
Ped variability & -11.288 & 1.778 & -6.349 & $<$0.001 (***) \\
Ped average density & -0.736 & 0.323 & -2.279 & 0.023 (*) \\
\hline
\end{tabular}}
\end{table}
For considering the effect of pedestrians' features all together, we fitted a generalized linear model (GLM, using maximum likelihood estimation) for predicting the label of clusters. All the features were centred to have a mean of zero for this analysis.
To avoid collinearity, pedestrian features that only had no or low correlation ($<0.3$) with each other were selected as fixed effects, and to improve model fit, the combination resulting in the least AIC was considered, which is reported in Table \ref{tab:glm}. The result of the GLM shows that all the three fixed effects of mean speed, variability, and average pedestrian density had a significant effect on the clustering result. Using other combinations of non-correlating features, other pedestrian features not considered in the model of Table~\ref{tab:glm} also had significant effects on predicting the clustering result. We also added the dataset as a random effect to the model (to take the effect of specific locations/datasets into account) and fit a generalized linear mixed-effect model (GLMM). This led to pedestrian path efficiency being the only significant fixed effect.

This difference may be because we only have one dataset representing cluster A, so, by including the dataset as a random effect, we are also accounting for the differences between the two environments. As we were not able to find more datasets representing an unstructured environment similar to SDD, future work is needed to further investigate the differences between the two environments.

\textbf{General Comparison of Environmental Features:} The environments in cluster A, overall are more dominated by pedestrians than vehicles as it can be seen from their lower vehicle over pedestrian ratios compared to cluster B (Fig.~\ref{fig:IbyC}). The higher stop fraction in pedestrians’ trajectory profile in cluster A, can be a usual behaviour observed on-campus where people might linger in the environment or regularly come to a stop along the path when meeting a friend. These behaviours will also cause high variability in the speed profile. The lower path efficiency of pedestrians in cluster A is equivalent to more deviations from a straight path which can be the trajectory behaviour of a pedestrian that is roaming around without having an exact destination point in mind. This behaviour is unlikely to happen for pedestrians that are crossing the road or are walking on the sidewalk of an urban area towards their destination.

Therefore, based on our results, We can point out some similarities between the trajectory features observed in cluster A and the characteristics of an unstructured environment. Slightly higher entropy of path orientations in cluster A (Fig. \ref{fig:PbyC}), is an indication of higher diversity in the path directions which happens when pedestrians are not restricted to follow designated paths under strict rules and are free to move anywhere. This was one of the main elements of the qualitative description of an unstructured environment.

\begin{figure*}
  \centering
  \includegraphics[width=0.9\linewidth]{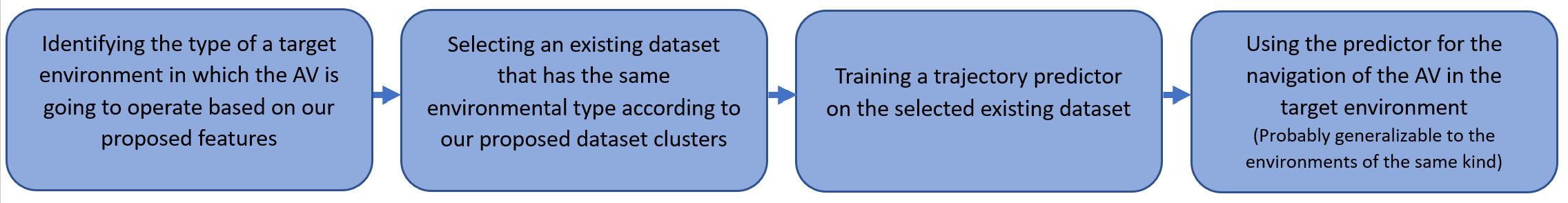}
  \caption{The process through which the results of this study can be used for developing trajectory prediction modules for autonomous vehicles (AVs)}
  \label{fig:Proc}
\end{figure*}

We can also describe an unstructured environment by the lower speed and higher variability of vehicles trajectory in those environments (Fig. \ref{fig:VbyC}), as there are no strict rules specifying the priorities or rights of way, and vehicles should pay more attention as pedestrians can be present everywhere. This could cause regular stops for the vehicles, resulting in slightly higher numbers for the stop fraction feature.

The effect of having no strict traffic rule in cluster A as an example of an unstructured environment can also be observed in the higher percentage of times that the vehicle yields and gives priority to pedestrians in the interactions happening in cluster A compared to cluster B (Fig. \ref{fig:IbyC}).
The pedestrian's dominance in the scenes of cluster A and their environment not having a road structure could have affected these priority behaviours in interaction which could be a common behaviour seen in unstructured environments. 

The number of standing pedestrians per unit area in cluster A is higher than in cluster B, showing that we should expect more standing pedestrians in environments similar to the ones in cluster A or in other words in unstructured environments. This is somehow also connected to having higher stop fraction within these environments.

Other features that could be added to the definition of an unstructured environment, suggested by our results, are moving pedestrians having slightly lower speed with a higher probability of coming to a stop along the path. This means that we should expect more variability in the pedestrians’ trajectory in an unstructured environment. These behavioural features are important to consider when predicting a pedestrian’s trajectory in different environments.

\section{Limitations and Future Work}
Our work had limitations. The number of existing datasets that include both pedestrians and vehicles is limited especially when it takes place outside the road structure, which is closer to the definition of an unstructured environment. An unstructured environment similar to the SDD dataset can be also found in other areas, such as in airport terminals and shopping malls, for which we could not find any data. Therefore, cluster A in our results only contains one major dataset (SDD) with its different scenes. Although having more datasets that could have fallen under cluster A would have strengthened our argument about the clusters representing unstructured vs. structured environments, there is still a clear difference between the SDD dataset and all the other datasets which are believed to be a result of the different nature of the environments. 

The other limitation comes from the differences between the field of view and the geometry of the environments where the data were captured from, with some datasets (e.g. DUT) being more focused on a small area as opposed to the others that have covered a wider region (e.g. in, SDD). This would affect the length of the trajectory for each agent present in the dataset with shorter trajectories probably showing fewer variations in the trajectory profile due to limited information available from them. Further analysis on a larger set of datasets from various environments in which both pedestrians and vehicles are present could help address these limitations. 

Although we tried to be comprehensive in the list of features extracted based on agents' trajectories, there could be other features that have not been considered which could reveal other differences between the two environment types. More features can be investigated in the future, as well as using a larger dataset that contains more examples of unstructured environments, which, to the best of our knowledge, is not presently available. Also, we propose investigating how a trajectory predictor trained on one dataset can be generalized to other environments of the same class. Fig~\ref{fig:Proc} shows how this work can be used in the process of developing a trajectory predictor for an autonomous vehicle in future work.

\section{Conclusion}

We studied the differences in the trajectory features of five existing datasets that contain both pedestrians and vehicles. After dividing the datasets into two clusters based on 13 different features related to agents' motion trajectories, we studied the effect of trajectory features on distinguishing between unstructured or structured nature of the captured environment. Among the resulting two clusters, one included mainly road-like environments which we named structured environment and the other class that contained scenes from a campus environment happening mainly outside of a road structure, was interpreted to be the unstructured environment class.  We discussed the differences between the motion behaviours of vehicles and agents in structured and unstructured environments and provided a quantitative measure for distinguishing between these environments. 
Considering these features, one can then use the related dataset for the similar type of environment to train a pedestrian trajectory predictor for learning the behavioural elements that are probable to be seen in the same type of environment.

\bibliographystyle{IEEEtran}
\bibliography{IEEEabrv,mybibfile}

\end{document}